\documentclass{article}
\usepackage{spconf,amsmath,graphicx,multicol}
\usepackage{amsfonts}
\usepackage{color}
\usepackage{bbm}
\usepackage{cite}
\usepackage{amssymb}
\usepackage{algorithm}
\usepackage{algorithmic}
\usepackage{tabularx}
\usepackage[utf8]{inputenc}
\usepackage[english]{babel}

\usepackage{amsthm}
\usepackage{subcaption}
\usepackage[font={footnotesize}]{caption}
\usepackage{url}
\usepackage{mathtools}
\usepackage{comment}
\usepackage{float}

\newtheorem{proposition}{Proposition}[]
\newtheorem{theorem}{Theorem}[]

\newcommand{\argmax}{\arg\!\max}

\DeclareMathOperator{\E}{\mathbb{E}}

\def\z{{\mathbf z}}
\def\n{{\mathbf n}}
\def\y{{\mathbf y}}

\def\C{{\mathbf C}}

\def\P{{\mathbf P}}
\def\R{{\mathbb{R}}}

\def\I{{\mathbf I}}

\def\x{{\mathbf x}}

\def\A{{\mathbf A}}
\def\B{{\mathbf B}}
\def\U{{\mathbf U}}
\def\V{{\mathbf V}}
\def\R{{\mathbb{R}}}

\def\I{{\mathbf I}}
\def\a{{\mathbf a}}

\def\u{{\mathbf u}}

\def\S{{\mathbf \Sigma}}

\newcommand{\ts}{\textsuperscript}

\let\oldref\ref
\renewcommand{\ref}[1]{(\oldref{#1})}
\newcommand{\RNum}[1]{\uppercase\expandafter{\romannumeral #1\relax}}
\makeatletter
\renewcommand{\fnum@figure}{Fig.~\thefigure}
\makeatother

\ninept
\title{Sampling and Reconstruction of Graph Signals via  Weak Submodularity and Semidefinite Relaxation}
\name{Abolfazl Hashemi$^\dagger$, Rasoul Shafipour$^\ddagger$, Haris Vikalo$^\dagger$, and Gonzalo Mateos$^\ddagger$}
\address{$^\dagger$Dept. of Electrical and Computer Engineering, University of Texas at Austin, Austin, TX, USA\\
$^\ddagger$Dept. of Electrical and Computer Engineering, University of Rochester, Rochester, NY, USA}
\begin{document}
%
\maketitle
\begin{abstract}
We study the problem of sampling a bandlimited graph signal in the presence of noise, where the objective is to 
select a node subset of prescribed cardinality that minimizes the signal reconstruction mean squared error (MSE). 
To that end, we formulate the task at hand as the minimization of MSE subject to binary constraints, and 
approximate the resulting NP-hard problem via semidefinite programming (SDP) relaxation. Moreover, we provide 
an alternative  formulation based on maximizing a monotone weak submodular function and propose a randomized-greedy 
algorithm to find a sub-optimal subset. We then derive a worst-case performance guarantee on the MSE returned 
by the randomized greedy algorithm for general non-stationary graph signals. The efficacy of the proposed 
methods is illustrated through numerical simulations on synthetic and real-world graphs. Notably, the randomized greedy 
algorithm yields an order-of-magnitude speedup over state-of-the-art greedy sampling schemes, while incurring 
only a marginal MSE performance loss.
\end{abstract}
\begin{keywords}
graph signal processing, sampling, weak submodularity, semidefinite programming, randomized algorithms \vspace{-0.2cm}
\end{keywords}
\section{Introduction}\label{sec:intro}
Consider a network represented by a graph $\mathcal{G}$ consisting of a node set $\mathcal{N}$ of 
cardinality $N$ and a weighted adjacency matrix $\mathbf{A} \in \mathbb{R}^{N \times N}$ whose
$(i,j)$ entry, $A_{ij}$, denotes weight of the edge connecting node $i$ to node $j$.  A \textit{graph 
signal} $\mathbf{x} \in \mathbb{R}^{N}$ can be viewed as a vertex-valued network 
process that can be represented by a vector of size $N$ supported on $\mathcal{N}$, where its $i$
\textsuperscript{th} component denotes the signal value at node $i$. Under the assumption that properties of the network process relate to the underlying graph, the goal of graph signal processing (GSP) is to generalize traditional signal processing tasks and develop algorithms that fruitfully exploit this relational structure \cite{shuman2013,sandryhaila2013}.

A keystone generalization which has drawn considerable attention in recent years pertains to sampling 
and reconstruction of graph signals \cite{shomorony2014sampling,tsitsvero2016signals,anis2016efficient,chen2015discrete,chepuri2016subsampling,marques2016sampling,gama2016rethinking,chamon2017greedy}. The task of finding an exact sampling set to perform reconstruction with minimal information loss is known to be NP-hard.
Conditions for exact reconstruction of graph signals from noiseless samples were put forth in~\cite{shomorony2014sampling,tsitsvero2016signals,anis2016efficient,chen2015discrete}. Sampling of noise-corrupted signals using randomized schemes including uniform and leverage score sampling is studied in \cite{chen2016signal}, for which optimal sampling distributions and
performance bounds are derived.
In \cite{chepuri2016subsampling,chamon2017greedy}, reconstruction of graph signals and their power spectrum density is studied and greedy schemes are developed. However, the performance guarantees 
in \cite{chen2016signal,chamon2017greedy} are restricted to the case of stationary graph signals, i.e., 
the covariance matrix 
in the nodal or spectral domains has certain structure (e.g., diagonal; see also \cite{marques2016stationaryTSP16,perraudinstationary2016,girault_stationarity}). 

In this paper, we study the problem of sampling and reconstruction of graph signals and propose two algorithms that solve it approximately. First, we develop a semidefinite programming (SDP) relaxation that finds a near-optimal sampling set in polynomial time. Then, we formulate the sampling task as that of maximizing a monotone weak submodular function and propose an efficient randomized greedy algorithm motivated by \cite{mirzasoleiman2014lazier}. We analyze the performance of the randomized greedy algorithm and in doing so, we show that the MSE associated with the selected sampling set is on expectation a constant factor away from that of the optimal set. Moreover, we prove that the randomized greedy algorithm achieves the derived approximation bound with high probability for every sampling task.  
In contrast to prior work, our results do not require stationarity of the signal. Finally, in simulation studies we illustrate the superiority of the proposed schemes over state-of-the-art randomized and greedy algorithms \cite{chen2016signal,chamon2017greedy} in terms of running time, accuracy, or both.
\footnote{MATLAB implementations of the proposed algorithms are available at \url{https://github.com/realabolfazl/GS-sampling}.} 


\noindent\textbf{Notation.} 
$\A_{ij}$ denotes the $(i,j)$ entry of matrix $\A$, $\a_j$ is the $j\ts{th}$ row of $\A$, $\A_{S,r}$ ($\A_{S,c}$) is a submatrix of $\A$ that contains rows (columns) indexed by the set $S$, and $\lambda_{\max}(\A)$
and $\lambda_{\min}(\A)$ represent the maximum and minimum eigenvalues of $\A$, respectively. $\I_N \in \R^{N\times N}$ denotes the identity matrix and $[N] := \{1,2,\dots,N\}$.
\vspace{-0.25cm}
\section{Preliminaries and Problem Statement}\label{sec:pre}
Let $\x$ be a zero-mean, random graph signal which is $k$-bandlimited in a given basis $\mathbf{V} \in \mathbb{R}^{N \times N}$. This means that the signal's so-called graph Fourier transform (GFT) $\bar{\x} = \V^\top \x$ is $k$-sparse. There are several choices for $\mathbf{V}$ in the literature with most aiming to decompose a graph signal into different modes of variation with respect to the graph topology. For instance, $\mathbf{V} = [\mathbf{v}_{1},\cdots,\mathbf{v}_{N}]$ can be defined via the Jordan decomposition of the adjacency matrix \cite{DSP_freq_analysis,deri2017spectral}, through the eigenvectors of the Laplacian when $\mathcal{G}$ is undirected \cite{shuman2013}, or it can be obtained as the result of an optimization procedure \cite{shafipour2017digraph,sardellitti}. We also assume that the signal in not necessarily stationary with respect to $\mathcal{G}$, and that $\bar{\x}$ is a zero-mean random vector with (generally non-diagonal) covariance matrix $\E[\bar{\x}\bar{\x}^\top]$ = $\P$. Recall that since $\x$ is bandlimited, $\bar{\x}$ is sparse with at most $k$ nonzero entries. Let $K$ be the support set of $\bar{\x}$, where $|K| = k$. Then, one can write $\x = \U\bar{\x}_K$, where $\U = \V_{K,c}$. 

Suppose that only a few (possibly noisy) entries of $\x$ can be observed, corresponding to 
taking measurements from a subset of nodes in $\mathcal{N}$. The goal of sampling is to
select the subset that enables reconstruction of the original signal with the smallest possible
distortion. Formally, let $\y = \x + \n$ be the noise-corrupted signal, where $\n \in \R^{N}$ is the zero-mean noise vector with covariance matrix $\E[\n\n^\top]=\sigma^2 \I_N$. Let $S \subseteq \mathcal{N}$ be a sampling subset of nodes and let $\hat{\x}$ be the reconstructed graph signal based on the measurements (i.e., the entries of $\y$) indexed by $S$. Since the signal is $k$-bandlimited, $\bar{\x}$ has at most $k$ nonzero entries. Therefore, we assume that the least mean square estimator of $\bar{\x}$ has at most  $k$ nonzero entries. This in turn imposes the constraint $|S|\leq k$. Now, 
since $\x = \U\bar{\x}_K$, the samples $\y_S$ and the non-zero frequency components of $\x$ are related via the Bayesian linear model
\begin{equation}\label{eq:model}
\y_S = \U_{S,r}\bar{\x}_K +\n_{S}.
\end{equation}
Hence, in order to find $\hat{\x}$ it suffices to estimate $\bar{\x}_K$ based on $\y_S$.
The least mean square estimator of $\bar{\x}_K$, denoted by $\hat{\bar{\x}}_{K_{\mathrm{lms}}}$, satisfies the Bayesian counterparts of the normal equations in the Gauss-Markov theorem (see e.g.,  \cite[Ch. 10]{kay1993fundamentals}). Accordingly, it is given by
\begin{equation}\label{eq:estf}
\hat{\bar{\x}}_{K_{\mathrm{lms}}} = \sigma^{-2}\bar{\S}_S \U_{S,r}^\top \y_S,
\end{equation}
where 
\begin{equation}\label{eq:covf}
\bar{\S}_S = \left(\P^{-1}+\sigma^{-2}\U_{S,r}^\top\U_{S,r}\right)^{-1}
\end{equation}
is the error covariance matrix of $\hat{\bar{\x}}_{K_{\mathrm{lms}}}$. Therefore, $\hat{\x} = \U\hat{\bar{\x}}_{K_{\mathrm{lms}}}$ and its error covariance matrix can be obtained as ${\S}_S = \U \bar{\S}_S \U^\top$.

The problem of sampling for near-optimal reconstruction can now be formulated as the task of choosing $S$ so as to minimize the mean square error (MSE) of the estimator $\hat{\x}$. Since the MSE is defined as the trace of the error covariance matrix, we obtain the following optimization problem,
\begin{equation}\label{eq:probs}
\begin{aligned}
& \underset{S}{\text{min}}
\quad \mathrm{Tr}\left({\S}_S\right)
& \text{s.t.}\quad S \subseteq \mathcal{N}, \phantom{k}|S|\leq k.
\end{aligned}
\end{equation}
Using trace properties and the fact that $\U^\top\U$ is a Hermitian positive semidefinite matrix, \ref{eq:probs} simplifies to
\begin{equation}\label{eq:probf}
\begin{aligned}
& \underset{S}{\text{min}}
\quad \mathrm{Tr}\left(\bar{\S}_S\right)
& \text{s.t.}\quad S \subseteq \mathcal{N}, \phantom{k}|S| \leq k.
\end{aligned}
\end{equation}
The optimization problem \ref{eq:probf} is NP-hard and evaluating all ${N}\choose{k}$ possibilities to find an exact solution makes it intractable even for relatively small graphs. In the next section we propose two alternatives to find near-optimal solutions in polynomial time.
\vspace{-0.2cm}
\section{New Schemes for Sampling Graph Signals}\label{sec:alg}
Here we resort to two approximation methods to find a near-optimal solution $S$ of \ref{eq:probf}. Our proposed algorithms are based on semidefinite and weak submodular optimization techniques that have recently shown superior performance in applications such as sensor selection\cite{joshi2009sensor}, graph sketching \cite{gama2016rethinking}, wireless sensor networks \cite{shamaiah2012greedy}, Kalman filtering \cite{ma}, and sparse signal recovery \cite{das2011submodular,hashemi2016sparse}. 
\vspace{-0.25cm}
\subsection{Sampling via SDP relaxation}
We first develop an SDP relaxation for problem \ref{eq:probf}. Our proposed scheme is motivated by the framework of \cite{joshi2009sensor} developed in the context of sensor scheduling. However, our focus is on sampling and reconstruction of graph signals which entails a different objective function, i.e., MSE.
Let $z_i \in \{0,1\}$ indicate whether the $i\ts{th}$ node of $\mathcal{N}$ is included in the sampling set $S$ and define $\z = [z_1,z_2,\dots,z_N]^\top$. Then, \ref{eq:covf} can alternatively be written as
\begin{equation}
\bar{\S}_z = \left(\P^{-1}+\sigma^{-2}\sum_{i=1}^n z_i\u_i\u_i^\top\right)^{-1}.
\end{equation}
Therefore, by relaxing the binary constraint $z_i \in \{0,1\}$ one can obtain a convex relaxation of  \ref{eq:probf},
\begin{equation}\label{eq:probf2}
\begin{aligned}
& \underset{\z}{\text{min}}
\quad \mathrm{Tr}\left(\bar{\S}_z\right)
& \text{s.t.}\quad 0\leq z_i \leq 1, \phantom{k} \sum_{i=1}^Nz_i \leq k.
\end{aligned}
\end{equation}
In order to obtain an SDP in standard form, let $\C$ be a positive semidefinite matrix such that $\C \succeq \bar{\S}_z$. 
Then, \ref{eq:probf2} is equivalent to 
\begin{equation}\label{eq:probf3}
\begin{aligned}
& \underset{\z,\C}{\text{min}}
\quad \mathrm{Tr}\left(\B\right)
& \text{s.t.}\quad 0\leq z_i \leq 1, \phantom{k} \sum_{i=1}^nz_i \leq k, \phantom{k} \C -\bar{\S}_z \succeq \mathbf{0}.
\end{aligned}
\end{equation}
The last constraint in \ref{eq:probf3}, i.e., $\C -\bar{\S}_z \succeq \mathbf{0}$, can be thought of as being the Schur complement \cite{horn2012matrix} of the block matrix
\begin{equation}
\B = \begin{bmatrix}
    \C& \I\\
    \I & \bar{\S}_z^{-1}
\end{bmatrix}.
\end{equation}
Note that the Schur complement of $\B$ is positive semidefinite if and only if $\B\succeq \mathbf{0}$ 
\cite{horn2012matrix}. Therefore, replacing the last constraint in \ref{eq:probf3} with the positive semidefiniteness 
constraint on $\B$ results in the following SDP relaxation:
\begin{equation}\label{eq:sdp}
\begin{aligned}
& \underset{\z,\B}{\text{min}}
\quad \mathrm{Tr}\left(\B\right)
& \text{s.t.}\quad 0\leq z_i \leq 1, \phantom{k} \sum_{i=1}^nz_i \leq k, \phantom{k} \B \succeq \mathbf{0}.
\end{aligned}
\end{equation}
An exact solution to \ref{eq:sdp} can be obtained by means of existing SDP solvers; see, e.g.,~\cite{arora2005fast,grant2008cvx}.  However, the solution $\hat{\z}$ contains real-valued entries and hence a rounding procedure is needed
to obtain a binary solution. Here, we propose to use the rounding procedure introduced in \cite{joshi2009sensor} and accordingly select the nodes of $\mathcal{N}$ corresponding to the $k$ $z_i$'s with largest values. The proposed SDP relaxation sampling scheme is summarized as Algorithm 1.
\renewcommand\algorithmicdo{}	
\begin{algorithm}[t]
\caption{SDP Relaxation for Graph Sampling}
\label{alg:sdp}
\begin{algorithmic}[1]
    \STATE \textbf{Input:}  $\P$, $\U$, $k$.
    \STATE \textbf{Output:} Subset $S\subseteq \mathcal{N} $ with $|S|=k$.
    \STATE Find $\z$, the minimizer of the SDP relaxation problem in \ref{eq:sdp} 
    \STATE Set $S$ to contain nodes corresponding to top $k$ entries of $\z$
	\RETURN $S$.
\end{algorithmic}
\end{algorithm}
\vspace{-0.25cm}
\subsection{Sampling via a randomized greedy scheme}
The computational complexity of the SDP approach developed in Section 3.1 might be challenging in applications 
dealing with large graphs. Hence, we now propose an iterative randomized greedy algorithm for the task of sampling 
and reconstruction of graph signals by formulating \ref{eq:probf} as the problem of maximizing a monotone weak 
submodular set function. First, we define the notion of monotonicity, submodularity, and curvature that will be useful 
in the subsequent analysis.

A set function $f:2^X\rightarrow \mathbb{R}$ is submodular if
\begin{equation}
f(S\cup \{j\})-f(S) \geq f(T\cup \{j\})-f(T)
\end{equation}
for all subsets $S\subseteq T\subset X$ and $j\in X\backslash T$. The term $f_j(S)=f(S\cup \{j\})-f(S)$ is the marginal value of adding element $j$ to set $S$. Moreover, the function is monotone if $f(S)\leq f(T)$ for all $S\subseteq T\subseteq X$. 

The concept of submodularity can be generalized by the notion of curvature or submodularity ratio \cite{das2011submodular} that quantifies how close a set function is to being submodular. Specifically, the maximum element-wise curvature of a monotone non-decreasing function $f$ is defined as
\begin{equation}
{\cal C}_{\max}=\max_{1\le l<n}{\max_{(S,T,i)\in \mathcal{X}_l}{f_i(T)\slash f_i(S)}},
\end{equation}
with $\mathcal{X}_l = \{(S,T,i)|S \subset T \subset X, i\in X \backslash T, |T\backslash S|=l,|X|=n\}$. Note that a set function is submodular if and only if ${\cal C}_{\max}~\le~1$. Set functions with ${\cal C}_{\max} > 1$ are called weak or approximate submodular functions \cite{das2011submodular}.

Next, similar to \cite{chamon2017greedy}, we formulate \ref{eq:probf} as a  set function maximization task. Let $f(S) = \mathrm{Tr}(\P-\bar{\S}_S)$. Then,  \ref{eq:probf} can equivalently be written as
\begin{equation}\label{eq:probsub}
\begin{aligned}
& \underset{S}{\text{max}}
\quad f(S)
& \text{s.t.}\quad S\subseteq\mathcal{N},\quad |S| \leq k.
\end{aligned}
\end{equation}

In Proposition \oldref{thm:p} below, by applying the matrix inversion lemma 
\cite{bellman1997introduction} we establish that $f(S)$ is monotone and weakly submodular. 
Moreover, we derive an efficient recursion to find the marginal gain of adding a new node 
to the sampling set $S$. 
\begin{proposition}\label{thm:p}
\textit{$f(S) = \mathrm{Tr}(\P-\bar{\S}_S)$ is a weak submodular, monotonically increasing set function, $f(\emptyset)=0$, and for all $j \in \mathcal{N}\backslash S$
\begin{equation}\label{eq:mg}
f(S\cup \{j\})-f(S) = \frac{\u_j^\top\bar{\S}_S^{2}\u_j}{\sigma^{2}+\u_j^\top\bar{\S}_S\u_j},\: \text{ and }
\end{equation}
\begin{equation}\label{eq:upf}
\bar{\S}_{S \cup\{j\}} = \bar{\S}_S-\frac{\bar{\S}_{S}\u_{j}\u_{j}^\top\bar{\S}_{S}}{\sigma^2+\u_{j}^\top\bar{\S}_{S}\u_{j}}.
\end{equation}}
\end{proposition}
Proposition \oldref{thm:p} enables efficient construction of the sampling set in an iterative fashion. To further reduce the computational cost, we propose a randomized greedy algorithm that performs the task of sampling set selection in the following way.  Starting with $S = \emptyset$, at iteration $(i+1)$ of the algorithm, a subset $R$ of size $s$ is sampled uniformly at random and without replacement from $\mathcal{N} \backslash S$. 
The marginal gain of each node in $R$ is found using \ref{eq:mg}, and the one corresponding to the highest marginal gain is added to $S$. Then, the algorithm employs the recursive relation \ref{eq:upf} to update $\bar{\S}_S$ for the subsequent iteration. This procedure is repeated until some stopping criteria, such as condition on the cardinality of $S$ is met. Regarding $s$, we follow the suggestion in \cite{mirzasoleiman2014lazier} and set $s=\frac{N}{k}\log\frac{1}{\epsilon}$, where $e^{-k}\leq \epsilon<1$ is a predetermined parameter that controls trade-off between the computational cost and MSE of the reconstructed signal; randomized greedy algorithm with smaller $\epsilon$ produces sampling solutions with lower MSE while the one with larger $\epsilon$ requires lower computational costs. Note that if $\epsilon = e^{-k}$, the randomized greedy algorithm in each iteration considers all the available nodes and hence  matches the greedy scheme proposed in \cite{chamon2017greedy}. However, as we illustrate in our simulation studies, the proposed randomized greedy algorithm is significantly faster than the method in \cite{chamon2017greedy} for larger $\epsilon$ while returning essentially the same sampling solution. The randomized greedy algorithm is formalized as Algorithm 2.

\renewcommand\algorithmicdo{}	
\begin{algorithm}[t]
\caption{Randomized Greedy Algorithm for Graph Sampling}
\label{alg:greedy}
\begin{algorithmic}[1]
    \STATE \textbf{Input:}  $\P$, $\U$, $k$, $\epsilon$.
    \STATE \textbf{Output:} Subset $S\subseteq \mathcal{N} $ with $|S|=k$.
    \STATE Initialize $S =  \emptyset$, $\bar{\S}_{S}=\P$.
	\WHILE{$|S|<k$}\vspace{0.1cm}
			\STATE Choose $R$ by sampling $s=\frac{N}{k}\log{(1/\epsilon)}$ indices uniformly at random from $\mathcal{N}\backslash S$
            \STATE $j_s = \argmax_{j\in R} \frac{\u_j^\top\bar{\S}_S^{2}\u_j}{\sigma^{2}+\u_j^\top\bar{\S}_S\u_j}$\vspace{0.1cm}
            \STATE $\bar{\S}_{S \cup\{j_s\}} = \bar{\S}_S-\frac{\bar{\S}_{S}\u_{j}\u_{j}^\top\bar{\S}_{S}}{\sigma^2+\u_{j}^\top\bar{\S}_{S}\u_{j}}$\vspace{0.1cm}
            \STATE Set $S \leftarrow S\cup \{j_s\}$\vspace{0.1cm}
	\ENDWHILE
	\RETURN $S$.
\end{algorithmic}
\end{algorithm}

\noindent{\textbf{Performance guarantees.}} Here we analyze performance of the randomized greedy algorithm. First, Theorem \oldref{thm:exp} below states that if $f(S)$ is characterized by a bounded maximum element-wise curvature, Algorithm 2 returns a sampling subset yielding an MSE that is on average within a multiplicative factor of the MSE associated with the optimal sampling set.
\begin{theorem}\label{thm:exp}
\textit{Let $\mathcal{C}_{max}$ be the maximum element-wise curvature of $f(S) = \mathrm{Tr}(\P-\bar{\S}_S)$, the objective function in problem \ref{eq:probsub}. Let $\alpha =(1-e^{-\frac{1}{c}}-\frac{\epsilon^\beta}{c})$, where $c=\max\{1,{\cal C}_{\max}\}$, $e^{-k}\leq\epsilon<1$, and $\beta = 1+\max\{0,\frac{s}{2N}-\frac{1}{2(N-s)}\}$. Let $S_{rg}$ be the sampling set returned by the randomized greedy algorithm and let $O$ denote the optimal solution of \ref{eq:probf}. Then,}
\begin{equation}\label{eq:expbound}
\E\left[\mathrm{Tr}(\bar{\S}_{S_{rg}})\right]\leq \alpha \mathrm{Tr}(\bar{\S}_{O}) + (1-\alpha) \mathrm{Tr}(\P).
\end{equation}
\end{theorem}
The proof of Theorem \oldref{thm:exp} relies on the argument that if $s = \frac{N}{k}\log\frac{1}{\epsilon}$, then with high probability the random subset $R$ in each iteration of Algorithm 2 contains at least one node from $O$.

Next, we study the performance of the randomized greedy algorithm using the tools of probably approximately correct (PAC) learning theory \cite{valiant1984theory,valiant2013probably}. The randomization of Algorithm 2 can be interpreted as approximating the marginal gains of the nodes selected by the greedy scheme proposed in \cite{chamon2017greedy}. More specifically, for the $i^{th}$ iteration it holds that $f_{j_{rg}}(S_{rg}) = \eta_i f_{j_{g}}(S_{g})$, where subscripts $rg$ and $g$ refer to the sampling sets and nodes selected by the randomized greedy (Algorithm 2) and greedy algorithm in \cite{chamon2017greedy}, respectively, and $0<\eta_i\leq 1$ for all $i \in [k]$ are random variables. 
In view of this argument and by employing the Bernstein inequality \cite{tropp2015introduction}, we obtain 
Theorem \oldref{thm:pac} which states that the randomized greedy algorithm selects a near-optimal sampling 
set with high probability.
\begin{theorem}\label{thm:pac}
\textit{Instate the notation and hypotheses of Theorem \oldref{thm:exp}. Assume $\{\eta_i\}_{i=1}^k$ are independent and let $C=0.088$. Then, with probability at least $1-e^{-Ck}$ it holds that}
\begin{equation}\label{eq:pacbound}
\mathrm{Tr}(\bar{\S}_{S_{rg}})\leq (1-e^{-\frac{1}{2c}}) \mathrm{Tr}(\bar{\S}_{O}) + e^{-\frac{1}{2c}} \mathrm{Tr}(\P).
\end{equation}
\end{theorem}
Indeed, in simulation studies (see Section \oldref{sec:sim}) we empirically verify the results of Theorems \oldref{thm:exp} and \oldref{thm:pac} and illustrate that Algorithm 2 performs favorably compared to the competing schemes both on average and for each individual sampling task. Before moving on to these numerical studies, 
in Theorem \oldref{thm:curv} we show that the maximum element-wise curvature of $f(S) = \mathrm{Tr}(\P-\bar{\S}_S)$ is bounded, even for non-stationary graph signals. 
\begin{theorem}\label{thm:curv}
\textit{Let $\mathcal{C}_{max}$ be the maximum element-wise curvature of $f(S) = \mathrm{Tr}(\P-\bar{\S}_S)$. Then, it holds that}
\begin{equation}\label{eq:curvbound}
\mathcal{C}_{\max} \leq \frac{\lambda_{\max}^2(\P)}{\lambda_{\min}^2(\P)}\left(1+\frac{\lambda_{\max}(\P)}{\sigma^2}\right)^3.
\end{equation}
\end{theorem}
An implication of Theorem \oldref{thm:curv} is a generalization of a result in \cite{chamon2017greedy} for stationary signals. There, it has been shown that if $\x$ is stationary and $\P = \sigma_\x^2\I_N$ for some $\sigma_\x^2>0$, then the curvature of the MSE objective is bounded. However, Theorem \oldref{thm:curv} holds even in the scenarios where the signal is non-stationary and $\P$ is non-diagonal.
\vspace{-0.55cm}
\section{Numerical Simulations}\label{sec:sim}
\begin{figure}[t]
	\vspace{-0.3cm}
	\begin{minipage}[b]{\linewidth}
		\centering
		\includegraphics[width=0.8\textwidth]{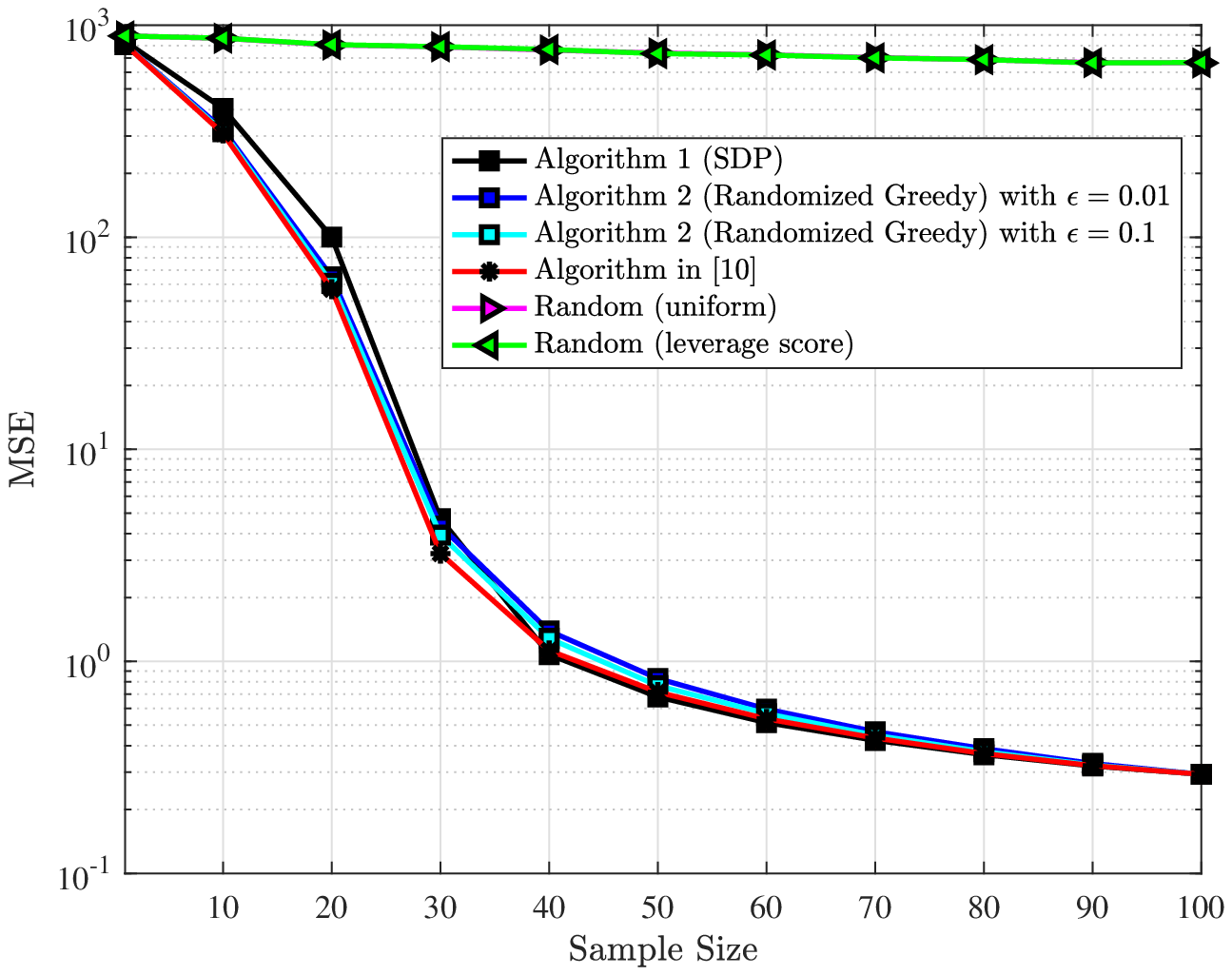}
	\end{minipage}
	\begin{minipage}[b]{\linewidth}
		\centering
		\includegraphics[width=0.8\textwidth]{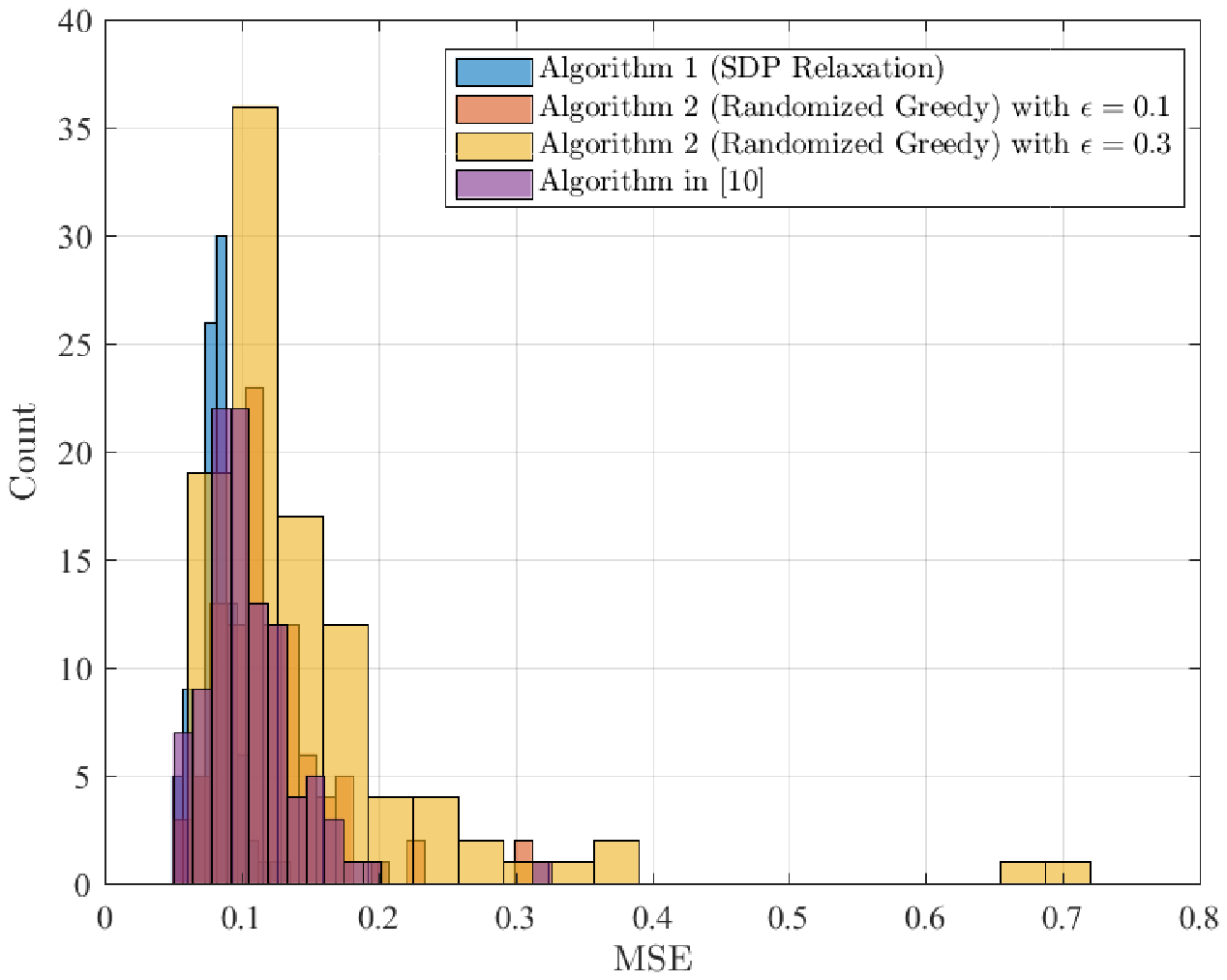}
	\end{minipage}
	\vspace{-0.5cm}
	\caption{Erd\H{o}s-R\'enyi graph. Comparison of different schemes in terms of (top) MSE as a function of the size of the sampling set; and (bottom) histogram of MSE values for $100$ realizations and fixed sampling set size. }
	\label{fig:rand}
	\vspace{-0.3cm}
\end{figure}

We study the recovery of simulated noisy signals supported on synthetic and real-world graphs to assess performance of the proposed sampling algorithms in terms of MSE and running time. To this end, we first 
consider an undirected Erd\H{o}s-R\'enyi random graph $\mathcal{G}$ of size $N=100$ and edge probability 0.2 \cite{newman2010networks}. Bandlimited graph signals $\x = \U\bar{\x}_K$ are generated by taking $\mathbf{U}$ as the first $k=30$ eigenvectors of the graph adjacency matrix. The non-zero frequency components $\bar{\x}_K$ are drawn from a zero-mean, multivariate Gaussian distribution with covariance matrix $\P$ which is selected uniformly at random from the set of positive semi-definite (PSD) matrices.  Zero-mean Gaussian noise $\n$ with covariance $\sigma^2 = 10^{-2}\I_N$ is added to $\x$. Algorithms \oldref{alg:sdp} and \oldref{alg:greedy} are run to recover the signal for different sampling set sizes. We compare the MSE performance of the proposed schemes with the state-of-the-art greedy algorithm \cite{chamon2017greedy} and the random sampling approaches in \cite{chen2016signal}. For the randomized greedy algorithm we use $\epsilon = 0.1$ and $\epsilon = 0.01$.  Fig. \oldref{fig:rand} (top) depicts the MSE versus $k$ (sample size), where the results are obtained 
by averaging over $100$ Monte-Carlo simulations. As the figure indicates, Algorithms \oldref{alg:sdp} and \oldref{alg:greedy} outperform the random sampling schemes of \cite{chen2016signal} and perform nearly as well as the greedy sampling algorithm \cite{chamon2017greedy}. While not shown here for the sake of clarity of the presentation, similar patterns were also observed for other workhorse random graphs, e.g., preferential attachment and Barb\'asi--Albert models \cite{newman2010networks}.

Next, we study the performance of the proposed schemes for each individual sampling tasks (each Monte-Carlo realizations),  for the setting where $N=10$ and $k=4$. 
Bandlimited graph signals are generated as before except that this time we take $\mathbf{U}$ as the first $4$ eigenvectors of the adjacency matrix. Fig. \oldref{fig:rand} (bottom) depicts superimposed MSE histograms of Algorithms \oldref{alg:sdp} and \oldref{alg:greedy} as well as the greedy sampling scheme \cite{chamon2017greedy} for 100 realizations per method and fixed $|S|=4$. As the figure illustrates, the proposed SDP relaxation and randomized greedy schemes perform well and are comparable with the greedy approach.

\begin{figure}[t]
	\vspace{-0.3cm}
	\begin{minipage}[b]{\linewidth}
		\centering
		\includegraphics[width=0.8\textwidth]{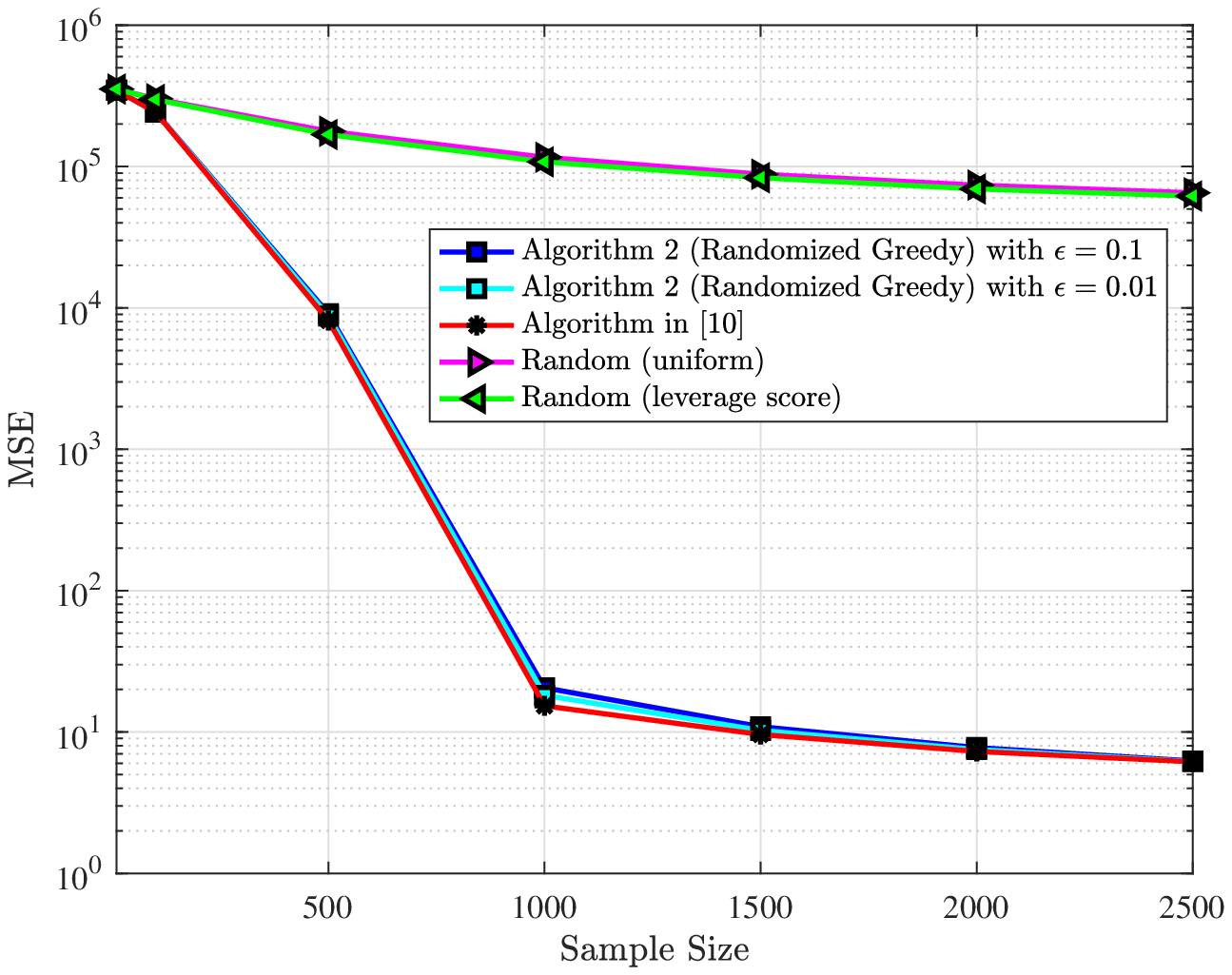}
	\end{minipage}
	\begin{minipage}[b]{\linewidth}
		\centering
		\includegraphics[width=0.8\textwidth]{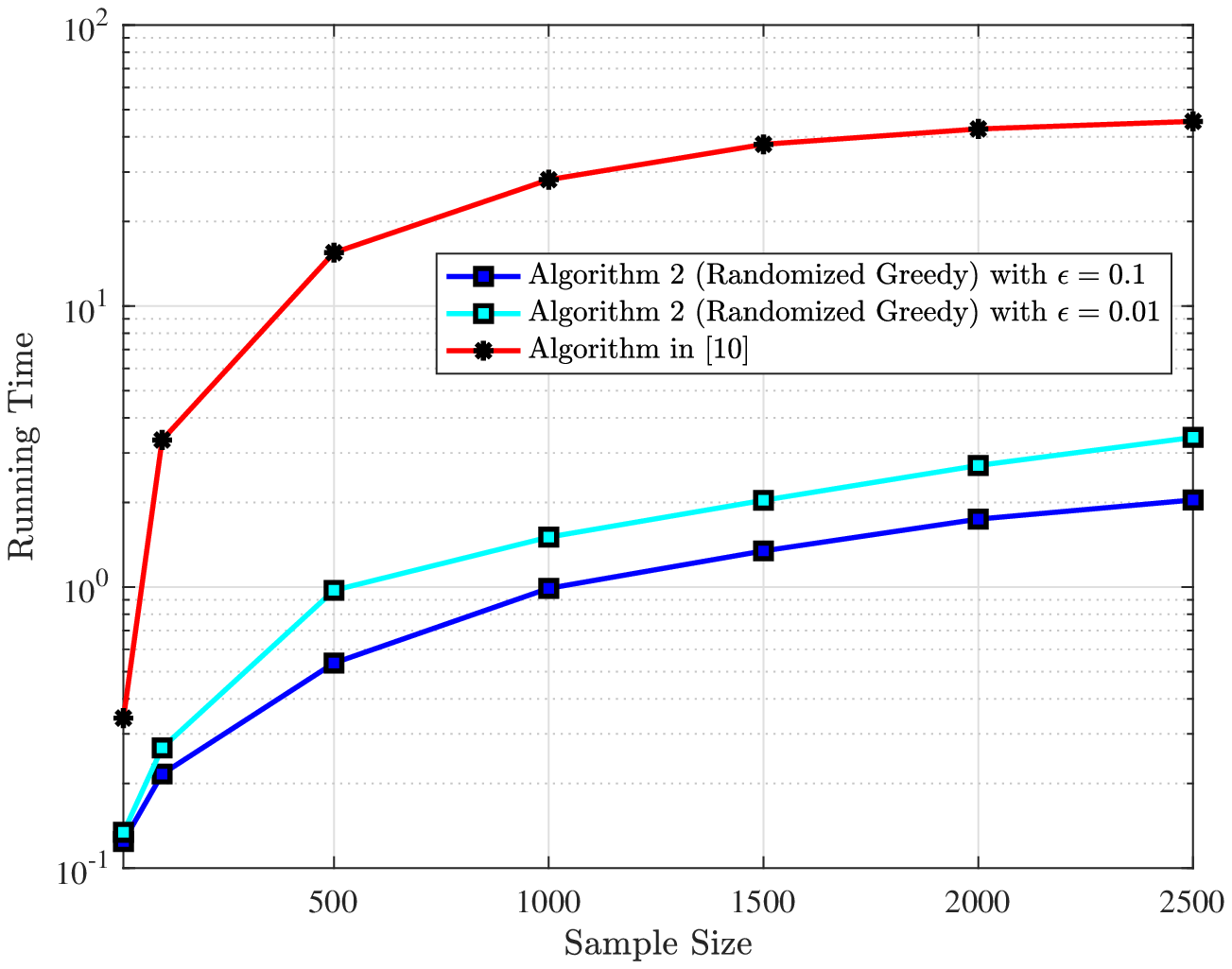}
	\end{minipage}
	\vspace{-0.5cm}	
	\caption{Minnesota road network. (top) MSE and (bottom) running time comparison of different sampling schemes as a function of the size of the sampling set.}
	\label{fig:min}
	\vspace{-0.3cm}
\end{figure}

Finally, we test Algorithm \oldref{alg:greedy} on the Minnesota road network\footnote{https://sparse.tamu.edu/Gleich/minnesota} with $N=2642$ nodes in order to showcase scalability of the proposed graph sampling method. To that end, Bandlimited graph signals are generated by taking the first $k=600$ eigenvectors of the graph Laplacian matrix, where the non-zero frequency components are drawn from a zero-mean, multivariate Gaussian distribution with randomly chosen PSD covariance matrix $\P$. The signals are corrupted with additive white Gaussian noise with $\sigma^{2}=10^{-2} \mathbf{I}_{N}$. 
As expected, Figs.~\oldref{fig:min} (top) and (bottom) depict trends of decreasing MSE and increasing running time versus $|S|$, respectively. The results are averaged over $1000$ Monte-Carlo simulations run on a commercial laptop with an Intel Core i$7$ processor at $3.1$ GHz. Remarkably, the proposed randomized greedy procedure achieves an order-of-magnitude speedup over the state-of-the-art algorithm in \cite{chamon2017greedy} while showing only a marginal degradation in the MSE performance.
\vspace{-0.2cm}
\section{Conclusion} \label{sec:concl}
We considered the problem of sampling a bandlimited graph signal in the presence of noise, where the goal is to select a subset of graph nodes of prescribed cardinality to minimize the mean square signal reconstruction error. First, we developed an SDP relaxation method to find an approximate solution to the NP-hard sample-set selection task. 
Then  the problem was reformulated as the maximization of a monotone weak submodular function, and a novel randomized-greedy algorithm was proposed to find a near-optimal sample subset. In addition, we analyzed the performance of the randomized greedy algorithm, and showed that the resulting MSE is a constant factor away from the optimal MSE. Unlike prior work, our guarantees do not require stationarity of the graph signal. Simulations studies showed that the proposed sampling algorithms compare favorably to competing alternatives in terms of accuracy and runtime.


\clearpage
\newpage
\bibliographystyle{IEEEbib}
\bibliography{refs}
\end{document}